\documentclass[journal]{IEEEtran}
\ifCLASSINFOpdf
\else
\fi

\usepackage{graphicx}
\usepackage{subfig}
\usepackage{epsfig}
\usepackage{amsmath}
\usepackage{enumerate}
\usepackage{subfig}

\usepackage[linesnumbered,ruled,vlined]{algorithm2e}
\usepackage{comment}
\usepackage{multirow}
\usepackage{dcolumn}
\newcolumntype{d}[1]{D{.}{.}{#1}}
\begin{document}
%

\title{ Relative Geometry-Aware Siamese Neural Network for 6DOF Camera Relocalization}
%
%
%
%

\author{Qing~Li,~\IEEEmembership{}
        Jiasong~Zhu,~\IEEEmembership{}
        Rui~Cao,~\IEEEmembership{}
        Ke~Sun,~\IEEEmembership{}
        Jonathan M. ~Garibaldi,~\IEEEmembership{}
        Qingquan~Li,~\IEEEmembership{}
        Bozhi~Liu,~\IEEEmembership{}
        and~Guoping~Qiu~\IEEEmembership{}

 }

\maketitle

\begin{abstract}
6DOF camera relocalization 
is an important component of autonomous driving and navigation. Deep learning has recently emerged as a promising technique to tackle this problem. 
In this paper, we present a novel relative geometry-aware Siamese neural network to enhance the performance of deep learning-based methods through explicitly exploiting the relative geometry constraints between images. We perform multi-task learning and predict the absolute and relative poses simultaneously. We regularize the shared-weight twin networks in both the pose and feature domains to ensure that the estimated poses are globally as well as locally correct. We employ metric learning and design a novel adaptive metric distance loss to learn a feature that is capable of distinguishing poses of visually similar images from different locations.
We evaluate the proposed method on public indoor and outdoor benchmarks and the experimental results demonstrate that our method can significantly improve localization performance. 
Furthermore, extensive ablation evaluations  are conducted to demonstrate the effectiveness of different terms of the loss function.
\end{abstract}


\IEEEdisplaynontitleabstractindextext

%
\IEEEpeerreviewmaketitle

\section{Introduction}\label{sec:introduction}

%
%
%
%
\IEEEPARstart{C}{amera} relocalization, or 6 degrees of freedom (6DOF) estimation, refers to the problem of estimating the pose (position and orientation) of an image (camera). It is a hot research topic in structure from motion (SfM), simultaneous localization and mapping (SLAM) and robotics, and it is also an essential component of autonomous driving and navigation.

Global Positioning System (GPS) has been widely used for vehicle localization but its accuracy significantly decreases in urban areas where tall buildings block or weaken its signals. Many image-based methods have been proposed to complement GPS. They provide position and orientation information based either on image retrieval \cite{murillo2009experiments,sattler2012image,ulrich2000appearance,wolf2005robust,wolf2002robust} or 3D model reconstruction \cite{kukelova2013real}. However, these methods face many challenges, including high storage overheads, low computational efficiency and image variations, especially for large scenes.

Recently, rapid progress in machine learning,  particularly deep learning, has produced a number of deep learning-based methods \cite{weyand2016planet,kendall2015posenet,szegedy2015going,kendall2015modelling,melekhov2017image,walch2017image,clark2017vidloc,kendall2017geometric,brahmbhatt2018geometry}. They have attained good performances in addressing the aforementioned challenges but their accuracies are not as good as traditional methods. 
Another severe problem of deep learning-based methods is that they fail to distinguish two different locations that have similar objects or scenes.
In this paper, we  present a novel relative geometry-aware Siamese neural network, which explicitly exploits the relative geometry constraints between images to regularize the network. We improve the localization accuracy and enhance the ability of the network to distinguish locations with similar images. It is achieved with three key new ideas:

\begin{enumerate}
\item We design a novel Siamese neural network that explicitly learns the global poses of a pair of images. We constrain the estimated global poses with the actual relative pose between the pair of images. 
\item We perform multi-task learning to estimate the absolute and relative poses simultaneously to ensure that the predicted poses are correct both globally and locally.

\item We employ metric learning and design an adaptive metric distance loss to learn feature representations that are capable of distinguishing the poses of similar visual images of different locations thus improving the overall pose estimation accuracy.

\end{enumerate}
%

The rest of the paper is organized as follows: Section \ref{sec:relatedwork} reviews the related works in camera relocalization. Section \ref{sec:deepRegression} elaborates the basic idea of deep learning-based camera relocalization methods.  Section \ref{sec:proposedmethod} describes the architecture of the proposed  network and its loss function items. We present the details of our experiments and evaluation in Section \ref{sec:experiment}. Finally, we conclude our work in Section \ref{sec:conclusion}.
\section{Related Work} \label{sec:relatedwork}

Camera relocalization methods can be mainly classified into three categories: image retrieval-based methods, 3D model-based methods, and deep learning-based methods.

Many approaches and systems are proposed based on image retrieval technique \cite{kalantidis2011viral,lee2015efficient,li2018geo,krose2001probabilistic,menegatti2004image,wang2006coarse,wang2005vision,cummins2008fab,torii2014efficient,umeda2018spherical,guzman2014multi}. They determine the pose of the query image  by matching it with images rendered from 3D scene models. The key component of the technique is image representation. Global descriptors are often used, such as colour histogram \cite{ulrich2000appearance} and gradient orientation histogram \cite{kosecka2003qualitative}. GIST descriptor \cite{oliva2001modeling} and GIST-based descriptors \cite{sunderhauf2011brief} are applied to represent panoramic images in \cite{singh2010visual,arroyo2014bidirectional,murillo2009experiments}. SeqSLAM \cite{milford2012seqslam} generates the global descriptor from a sequence of consecutive images instead of a single image. Global descriptors are fast to compute, but they are not robust to occlusion and illumination changes. Local features like SIFT \cite{lowe1999object} and SURF \cite{bay2008speeded}, have been used in \cite{lee2015efficient} for image representation. Compared with the global descriptor, they are less sensitive to occlusion and view variations. However, the storage requirement of the method is high for large scenes. The pooling features like BoW \cite{sivic2003video} and VLAD \cite{jegou2010aggregating} are able to relieve the challenge. They aggregate local  features and represent the locations with a compact feature vector instead of a large number of local features \cite{kalantidis2011viral}.

Another type of methods solve the problem by utilizing camera projection geometry between 2D pixels and 3D models. They estimate the pose by constructing the correspondence between 2D pixels and 3D points of the scene \cite{donoser2014discriminative,sattler2015hyperpoints,sattler2017efficient,sattler2012improving,uyttendaele2012real}. Local point features, like SIFT \cite{lowe1999object}, SURF \cite{bay2008speeded} and ORB \cite{rublee2011orb}, are frequently used to describe the detected 2D points. 3D points,  generated using the SfM technique, are also described with local features to perform 2D-3D matching. It can achieve accurate results when enough correct pairs are provided. The main challenge is to establish enough correct 2D-3D correspondences, which is difficult for two reasons. Firstly, local feature descriptor fails when a scene has repetitive texture or texture-less surface; 
and secondly, the process is  inefficient for large scenes.

To increase the efficiency of the 2D-3D matching, prioritized search approaches \cite{sattler2017efficient,sattler2012improving} are proposed to construct enough matching pairs instead of matching all detected 2D points. Scene coordinate random forest (SCRF) \cite{shotton2013scene, valentin2015exploiting} utilizes machine learning techniques to directly predict 3D coordinates of image pixels by training a random forest. Similar to SCRF, deep learning technique is employed to predict 3D coordinate of the center point of an image patch in \cite{brachmann2017dsac}. However, these methods require 3D model for the network training ,which limits their application. To filter out the wrong matches, co-visibility information is exploited in \cite{sattler2015hyperpoints,sattler2017efficient}.

Deep learning has achieved extraordinary performance in image classification, object detection, and image retrieval tasks. 
Many researchers have employed it to solve the camera relocalization problem \cite{weyand2016planet,kendall2015posenet,szegedy2015going,kendall2015modelling,melekhov2017image,walch2017image,clark2017vidloc,kendall2017geometric,brahmbhatt2018geometry}. PlaNet\cite{weyand2016planet} regards the problem as a classification task. It divides the map into grids and predicts the grid in which the query image belongs to through deep learning technique.  Many other researchers consider it as a regression problem instead. They directly estimate the pose through a convolutional neural network. PoseNet \cite{kendall2015posenet}, built on the GoogLeNet model \cite{szegedy2015going}, is the first to adopt this paradigm in an end-to-end manner.  It is further extended to Bayesian PoseNet \cite{kendall2015modelling} to estimate the confidence of the result as well. HourglassNet \cite{melekhov2017image} utilizes the encoder-decoder network structure with skipped connections to aggregate features from both lower and higher layers for pose regression. It achieves better performance than PoseNet. LSTM-Net \cite{walch2017image} believes that high dimensional output of fully connected layer in PoseNet is not optimal. It adds a LSTM network after the last fully connected layer in PoseNet to reduce information redundancy. VidLoc \cite{clark2017vidloc} exploits smooth constraints of a video to address the perceptual aliasing problem. It takes a video clip as input instead of a single image and proposes a bidirectional recurrent neural network structure to fuse the previous and next images information to increase predicted pose accuracy. Laskar \cite{laskar2017camera} proposes a new triangulating strategy that predicts the pose by estimating the relative pose between the query image and the images in the database. Its main drawback is low efficiency since  the relative pose of all the images in the database have to be computed. PoseNet2 \cite{kendall2017geometric} introduces the re-projection error with global pose error and improves the performance. However, 3D points are required in their method. MapNet \cite{brahmbhatt2018geometry} fuses the inertial information with image information through deep learning to enhance the network performance.

The proposed method in this paper is also based on convolutional neural networks. However, it has a number of distinctive features. For example, we use an innovative Siamese network architecture to exploit the relative geometry of images in addition to predicting the absolute poses. 
Unlike \cite{kendall2017geometric} and \cite{brahmbhatt2018geometry}, we only rely on  the 2D images for training. Compared to \cite{kendall2015modelling,kendall2015modelling,melekhov2017image,walch2017image}, we take a pair of images as input and utilize their relative pose error for training. In contrast to \cite{laskar2017camera}, we directly regress the image pose instead of performing triangulation.

A very recent work that also uses multi-task learning and explicitly models relative poses of two frames appears in \cite{valada2018deep, radwan2018vlocnet++}. However, our system architecture differs from that of  \cite{valada2018deep, radwan2018vlocnet++} in a number of significant ways.  Whilst we use a Siamese network and metric learning loss to model the relative geometrics of two frames, \cite{valada2018deep, radwan2018vlocnet++} use two separate networks to model the relative geometrics of two consecutive frames (although  \cite{valada2018deep, radwan2018vlocnet++} refer their two networks as Siamese network, strictly speaking it is not a Siamese network architecture because the two networks do not share weights). Furthermore, while our method can model the relative geometrics of two arbitrary frames,  \cite{valada2018deep, radwan2018vlocnet++} can only model two consecutive frames.
\section{Deep Learning-based Camera Relocalization} \label{sec:deepRegression}
Deep learning-based camera relocalization methods use an end-to-end learning strategy to predict the positions and orientations directly. They do not  perform image matching or solve 2D-3D correspondence as traditional methods do. Instead, they regard the task as a regression problem and utilize convolutional neural networks to model the hidden mapping function between the images and their corresponding poses. The networks are supervised by the distance between the predicted poses and the ground truth. This section focuses on discussing the pose representation  and describing the loss function formulation.
\subsection{Pose Representation}
The image (camera)  pose is comprised of the positional component and the orientational component. The position is denoted by a 3-dimensional vector {\bfseries x} of the arbitrary coordinate space. Orientation can be represented in 3 forms: Euler angle, transformation matrix, and quaternion. Euler angle is not a good choice because it suffers from the gimbal lock problem. Transformation matrix is over-parameterized for orientation because it contains 9 parameters to represent the orientation of 3D space, while the orientation only has 3 degrees of freedom. Previous works \cite{kendall2015posenet,clark2017vidloc,melekhov2017image,walch2017image} choose the quaternion to represent orientation, because it is a smooth and continuous representation. The quaternion is a 4-dimensional unit vector {\bfseries q} and is easy to perform back-propagation. The main concern for the quaternion is that each orientation has two different quaternion representations. This can be addressed by constraining the quaternion to one hemisphere.

One simple and obvious way to represent pose is to form a 7-dimensional vector, combining position and orientation together.
However, previous works demonstrate that the 7-dimensional vector representation does not achieve good performance due to the difference of scale between position and orientation. Therefore, two pose components are usually regressed separately. In this paper, instead of training two separate convolutional neural networks to estimate position and orientation, we train one model to predict the two components simultaneously. This is reasonable because both position and orientation come from the same image content.

\subsection{Loss Function}
The loss function (GlobalLoss) is normally designed based on the distance between the predicted pose and the ground truth, serving as the optimization objective for training the networks. It consists of two components, i.e. position loss and orientation loss, as shown in equation (\ref{loss:global_overall}).
\begin{equation}\label{loss:global_overall}
  L_{G} = L_{Gx}+ L_{Gq},
\end{equation}
where $L_{Gx}$ is the  position loss and $L_{Gq}$ denotes the  orientation loss.
Here, Euclidean distance is chosen to calculate the position loss and orientation loss as it is continuous and smooth. The two components are computed by equations \eqref{global_pos_loss} and  \eqref{global_ort_loss} respectively.
\begin{equation}\label{global_pos_loss}
  L_{Gx} = \left\|x-\hat{x}\right\|_{2},
\end{equation}
where x represents the real position and $\hat{x}$ denotes the predicted one.

\begin{equation}\label{global_ort_loss}
  L_{Gq} = \left\|q-\frac{\hat{q}}{\left\|\hat{q}\right\|}\right\|_{2},
\end{equation}
where $q$ is the ground truth  orientation, $\hat{q}$ denotes the predicted orientation and $||\hat{q}||$ represents the length of the predicted orientation quaternion. $\frac{\hat{q}}{\left\|\hat{q}\right\|}$ is performed to normalize the predicted quaternion to the length of 1 since the network prediction does not guarantee it.

Due to the quantity and scale difference between the position loss and the orientation loss, a  hyperplane  parameter $\beta$ is introduced to balance the influence of the two loss components. The loss function is represented as equation \eqref{loss:beta}.
\begin{equation}\label{loss:beta}
  L = L_{Gx}+\beta \times L_{Gq},
\end{equation}
Previous works choose to set $\beta$ manually and achieve good performance in their experiments. However, fine tuning $\beta$ for different scenes is labour-intensive. PoseNet2 addresses this issue by introducing two learnable variables, i.e. $\hat{s}_{x}$ and $\hat{s}_{q}$, which correspond to the loss of position and orientation respectively. Then equation \eqref{loss:beta} is transformed into equation \eqref{final_Globalloss}:

\begin{equation}\label{final_Globalloss}
  L = L_{Gx} \times exp(-\hat{s}_{x})+\hat{s}_{x}+ L_{Gq} \times exp(-\hat{s}_{q})+\hat{s}_{q}.
\end{equation}

\section{Relative Geometry-Aware Siamese  Network for Camera Relocalization} \label{sec:proposedmethod}
\begin{figure*}[!t]
\centering
\includegraphics[width=120mm]{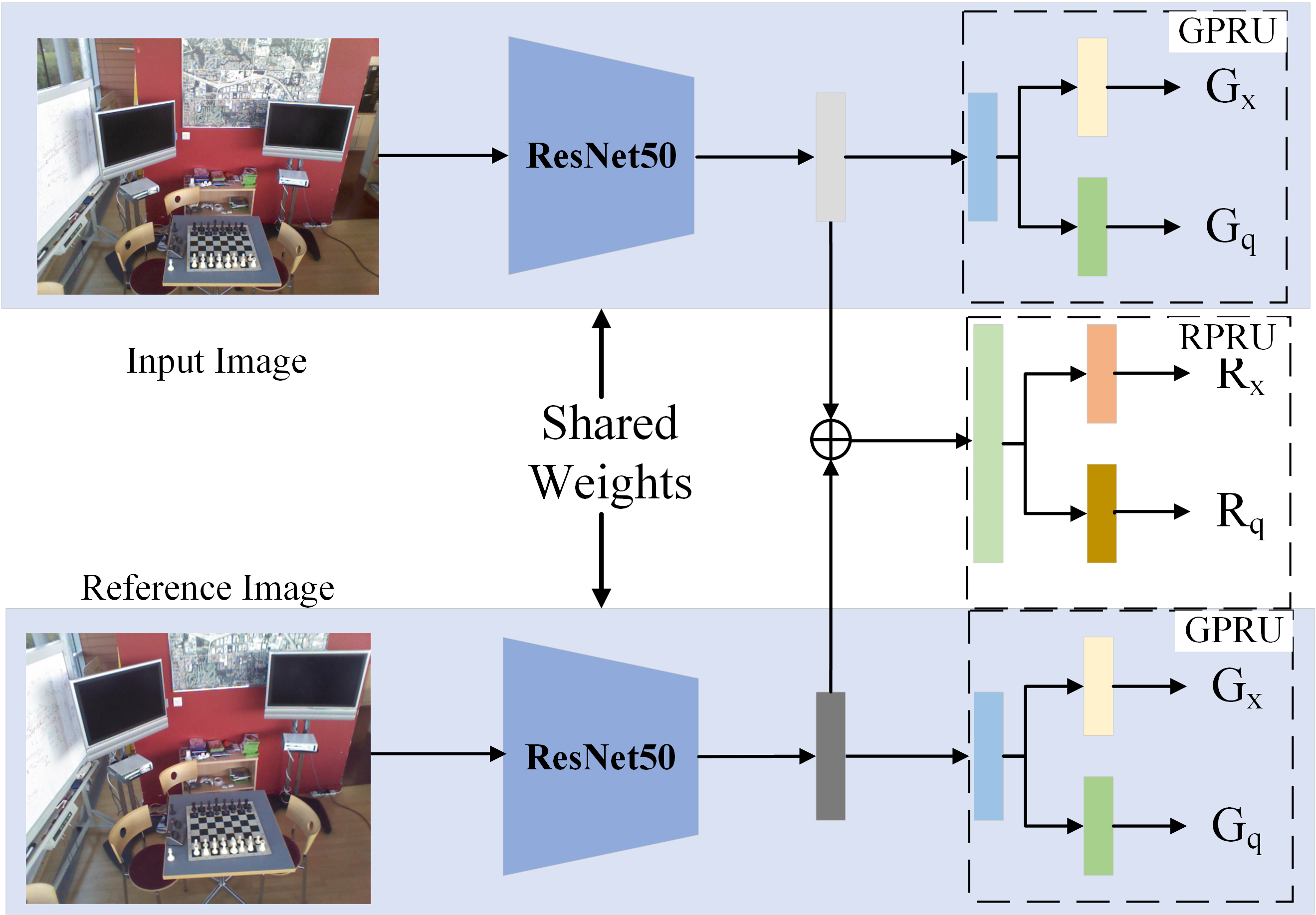}
\caption{Relative Geometry-Aware Siamese neural network architecture for 6DOF camera relocalization. Units of the same color share the same weights. The silver and grey unit represent the outputs of the modified ResNet50. $G_{x}, G_{q}$ denote the positional and orientational components of the predicted global pose, and $R_{x},R_{q}$ denote two components of the predicted relative pose. The global pose regression unit (GPRU) and the relative pose regression unit (RPRU) are represented with dashed-boundary boxes.}\label{fig:NetworkStructure}
\end{figure*}

Our network is built on Siamese network originally introduced by Bromley and LeCun in \cite{bromley1994signature}. A traditional Siamese neural network architecture consists of twin networks which accepts distinct inputs. The loss function computes a metric between the highest-level feature representation on each side given certain threshold. We utilize this structure to learn a robust feature representation for mapping positions and orientations by introducing relative geometry constraints of the training images. The process is supervised by both global pose and relative pose constraints. The proposed network architecture is illustrated in Figure \ref{fig:NetworkStructure}. Compared to the conventional Siamese network structure, it has an additional component for relative pose prediction and performs multi-task learning. In the following subsections, we will present the network architecture  and the relative geometry losses for  the network training in detail.


\subsection{ Network Architecture}
Each of the twin networks consists of a modified ResNet50 \cite{he2016deep} and a global pose regression unit (GPRU). The modified ResNet50 consists of 5 residual blocks and an average pooling layer. Each residual block has multiple residual bottleneck units that are comprised of three convolutional layers with kernel sizes of $1 \times 1$, $3 \times 3$, and $1 \times 1$ in sequence. Each convolutional layer is followed by rectified linear unit (ReLU)  and batch normalization operation. The average pooling layer is used to aggregate the feature information from the previous layers. The GPRU contains 3 fully connected layers. The first fully connected layer has 1024 neurons and the followed two has 3 and 4 neurons respectively for regressing the position and orientation. For the relative pose of the two inputs, we design a relative pose regression unit (RPRU). It has a similar structure as the GPRU. The difference lies in their inputs. While the GPRU takes the output vector of the modified ResNet50 as input, the RPRU takes the concatenation of the two modified ResNet50 output vectors as input. The dropout technique is applied after each fully connected layer to reduce feature redundancy. The parameter of dropout layer is set to be 0.2 empirically.

\subsection{Relative Geometry Losses}
We design three relative geometry losses based on the relative geometry constraints of the training images including the relative pose loss (RelLoss), the relative pose regression loss (RelRLoss) and the adaptive metric distance loss (MDLoss). They function in $both$ the feature and the pose spaces to regularize the network. They will be discussed in detail in the following sections.


\subsubsection{Relative Pose Loss}\label{subsubsec_RPCL}

Previous deep learning-based pose estimation methods train the network  on the global poses of the images, i.e. given an input image, they estimate its global (absolute) position and orientation while the relative pose between two training images is ignored. However, the relative pose information of two images is important. In this paper, the network not only explicitly estimates the global pose of the input image but also explicitly requires that the difference between the estimated global poses of two images is consistent with their actual (ground truth) difference. The relative pose loss (RelLoss) is designed to preserve the relative geometry in the pose space by comparing the distance between two predicted global poses, and the actual distance of the global poses of the two images. RelLoss is able to keep the relative pose of paired images consistent with their ground truth. It works in the pose space and constrains the pose error of two images.

Suppose that the position and orientation of the current image $I$ and a reference image $I_{ref}$ are ($x$, $q$) and ($x_{ref}$, $q_{ref}$), respectively. The relative position $x_{rel}$ and orientation $q_{rel}$  can be computed with equations \eqref{e:rel_pos} and \eqref{e:rel_ort}.
\begin{equation}\label{e:rel_pos}
  x_{rel} = x - x_{ref},
\end{equation}

\begin{equation}\label{e:rel_ort}
  q_{rel} = q^{*}_{ref} \times q,
\end{equation}
where $q^{*}_{ref}$ represents the conjugate quaternion of $q_{ref}$.
Note that when calculating the relative orientation from the predicted orientation quaternion with equation \eqref{e:rel_ort}, the quaternion has to be normalized. The RelLoss also contains the positional loss component and the orientational loss component as shown in equation \eqref{loss:relativeConstancy}.

\begin{equation}\label{loss:relativeConstancy}
  L_{C} = L_{Cx}+L_{Cq},
\end{equation}
where $L_{Cx}$ denotes the RelLoss positional component, and $L_{Cq}$ is the orientational component.

The two loss components are formulated with Euclidean distance as shown in equations  \eqref{loss:relativeConstancy_pos} and \eqref{loss:relativeConstancy_ort}.
\begin{equation}\label{loss:relativeConstancy_pos}
  L_{Cx} = \left\|\hat{x}_{rel}-x_{rel}\right\|_{2},
\end{equation}

\begin{equation}\label{loss:relativeConstancy_ort}
  L_{Cq} = \left\|\hat{q}_{rel}-q_{rel}\right\|_{2},
\end{equation}
where $\hat{x}_{rel},\hat{q}_{rel}$ are the predicted relative position and orientation, and $x_{rel},q_{rel}$ denote the ground truth.

\subsubsection{Relative Pose Regression Loss}
Whilst RelLoss captures the relative geometry of two images through estimating their global poses, we here introduce another loss to estimate the relative pose distance of a pair of images directly from the input images.  
The relative pose regression loss (RelRLoss) is defined as shown in equation  \eqref{loss:relative regression loss}.
\begin{equation}\label{loss:relative regression loss}
  L_{R} = L_{Rx}+L_{Rq},
\end{equation}
where $L_{Rx}$ denotes the  positional component, and $L_{Rq}$ denotes  the orientational component.
The two component loss functions are computed by equations \eqref{direc_rel_pos_loss} and \eqref{direc_rel_ort_loss}.

\begin{equation}\label{direc_rel_pos_loss}
  L_{Rx} = \left\|x_{rel}-\widetilde {x}_{rel}\right\|_{2},
\end{equation}

\begin{equation}\label{direc_rel_ort_loss}
  L_{Rq} = \left\|q_{rel}-\frac{\widetilde {q}_{rel}}{\left\|\widetilde {q}_{rel}\right\|}\right\|_{2},
\end{equation}

where $x_{rel}, q_{rel} $ represent the ground truth relative position and orientation, and $\widetilde{x}_{rel},\widetilde{q}_{rel}$ represent the directly predicted relative position and orientation. The ground truth relative position and orientation can be obtained using equation \eqref{e:rel_pos} , \eqref{e:rel_ort}. Note that $\widetilde{q}_{rel}$ needs to be normalized as it is directly regressed by the network.

It should be noted that $L_{R}$ in equation \eqref{loss:relative regression loss} and $L_{C}$ in equation \eqref{loss:relativeConstancy} are different. One is computed from the difference of two predicted global poses while the other is predicted directly by regression. Furthermore, it is the $L_{R}$ that joins the twin networks together (please refer to Figure \ref{fig:NetworkStructure}). The purposes of introducing RelRLoss is to ensure that the features extracted by the ResNet50 network will not only enable an accurate estimate of the global pose but also an accurate relative pose estimation.

\subsubsection{Adaptive Metric Distance Loss}
Deep learning-based methods often fail to accurately predict the poses of similar images of different locations. Distinguishing similar inputs belonging to different classes is one of the major difficulties in computer vision. Here, we take advantage of the Siamese network architecture of Figure \ref{fig:NetworkStructure} and propose the adaptive metric distance loss (MDLoss) to address the problem. It is inspired by metric learning \cite{bellet2013survey,norouzi2012hamming,chopra2005learning}. The basic idea of metric learning is to learn a metric distance adaptive to the problem of interest. For many problems, including camera relocalization, hand-crafted representations fail badly in capturing the notion of similarity. Deep learning regression-based camera relocalization approaches are based on the visual contents of the input image to estimate its pose, therefore simple metrics measuring the visual content similarity fails to capture the pose dissimilarity in the above cases.
In the case of our Siamese architecture in Figure \ref{fig:NetworkStructure}, the 6DOF camera pose is estimated by the GPRU. The input to the GPRU unit (the output of the ResNet50) should reflect the \textit{pose difference} rather than the \textit{visual similarity} of the images. We therefore introduce the adaptive metric distance loss (MDLoss) to address this issue.

The MDLoss is built on the contrastive loss, which employs semantic information (data label) to force the convolutional neural network to learn an embedding representation that complies with a notion of similarity of the problem domain. In our scenario, we define the metric distance loss by embedding the relative pose of two images. The relative information is used to define the margin of feature representation. The loss function is shown in equation \eqref{e:metric_distance_loss}.
\begin{equation}\label{e:metric_distance_loss}
L_{MD} = \frac{1}{2N}\sum_{n=1}^{N}\{\max(d_{x}+\alpha \times d_{q} - d,0)\}^2,
\end{equation}
where $N$ denotes the number of the training samples, $d = ||f-f_{ref}||_{2}$,  $f$ and $f_{ref}$ are the outputs of the modified ResNet50 network taken for the current image and the reference image respectively, $d_{x} = ||x-x_{ref}||_{2}$ is the Euclidean distance of the actual relative position while $d_{q} = ||q-q_{ref}||_{2}$ is the Euclidean distance of the actual relative orientation of the current image and the reference image, $\alpha$ is a positive constant to balance the influence of the relative position and orientation. It is set equal to 10 empirically.

An explanation of $L_{MD}$ is that, if $d $ is smaller than $d_{x}+\alpha \times d_{q}$, we want to make it as large as $d_{x}+\alpha \times d_{q}$. On the other hand, if $d $ is larger than $d_{x}+\alpha \times d_{q}$, this cost function is not utilized and other cost functions will function to ensure $f$ and $f_{ref}$ to take the appropriate values. This is a reasonable strategy because the reference image is always taken at a different location from that of the current image.





\subsection{Comprehensive Loss}
We train the proposed neural network jointly with GlobalLoss, RelLoss, RelRLoss and MDLoss. The comprehensive loss can be represented by equation \eqref{init_overall_loss}.
\begin{equation}\label{init_overall_loss}
  L = L_{G}+L_{C}+L_{R}+L_{MD}
\end{equation}

Equation \eqref{init_overall_loss} can also be written in the form of equation \eqref{loss:trains}.
\begin{equation}\label{loss:trains}
  L = L_{x} +L_{q} + L_{MD},
\end{equation}

It consists of three components: position loss $L_{x}$, orientation loss $L_{q}$ and metric distance loss $L_{MD}$. Positional loss  and orientational loss each has three components and can be written as equations \eqref{loss_pos} and \eqref{loss_ort} respectively.

\begin{equation}\label{loss_pos}
  L_{x} = L_{Gx}+L_{Cx}+L_{Rx},
\end{equation}

\begin{equation}\label{loss_ort}
  L_{q} = L_{Gq}+L_{Cq}+L_{Rq}.
\end{equation}

We choose a learning strategy to balance the position loss $L_{x}$ and orientation loss $L_{q}$ similar to PoseNet2. Therefore, the comprehensive loss can be further reformulated as equation  \eqref{final_loss}:

\begin{equation}\label{final_loss}
  L = L_{x} \times \exp(-\hat{s}_{x})+\hat{s}_{x} +L_{q} \times \exp(-\hat{s}_{q})+\hat{s}_{q} + L_{MD}.
\end{equation}
where $\hat{s}_{x}$ and $\hat{s}_{q}$ are learnable coefficients. 

\section{Experiments} \label{sec:experiment}
In this section, we test our method on two publicly available camera relocalization  benchmark datasets, one indoor and one outdoor, to demonstrate its effectiveness. Experimental results are presented and compared with  state-of-the-art methods in the literatures. 
We also investigate the role of various components of the loss function and analyze how the choice of reference image affects the performance of the proposed method.

\subsection{Datasets}
The two public datasets we used are: \emph{7Scene} \cite{shotton2013scene} and \emph{Cambridge Landmarks} \cite{kendall2015posenet}. To make our results exactly comparable to previous methods, we use the same split of training set and testing set as in the original datasets. 

\emph{7Scene} is an indoor image dataset for camera relocalization and trajectory tracking. It is collected with a handhold RGB-D camera. The ground truth pose is generated using the Kinect Fusion approach \cite{newcombe2011kinectfusion}. The dataset is captured in 7 indoor scenes. For each scene, it contains several image sequences, which has already been divided into training and testing sets. The images are taken at the resolution of  $640\times480$ with known focal length of 585. The dataset is quite challenging as motion makes the images blur. Besides, the indoor scenes are usually texture-less, which makes the localization problem even more difficult.

\emph{Cambridge Landmarks} is an outdoor dataset collected in 4 sites around Cambridge University. It is collected using a Google mobile phone while pedestrians walk. The images are captured at the resolution of $1920\times1080 $ and the ground truth pose is obtained through VisualSFM software \cite{wu2011visualsfm}. The dataset is also very challenging as it is taken in different weather and lighting conditions. Besides, the occlusion of moving pedestrians and vehicles further increases the difficulty.

\subsection{Setup}
\emph{Training phase}: in this phase, all parts of the proposed network are involved. It takes in a pair of images and outputs the corresponding global poses of them. It is important to note that, the twin networks are \textit{identical}. One takes the current image as input and produces its global 6DOF pose information, while the other takes the reference image as input and outputs its corresponding pose.

\emph{Testing phase}: in the testing phase, only \textit{one} of the twins is necessary. Since they are identical, any one can be used. The middle part that linking the twins is no longer necessary in this stage. Once training is completed, an image is fed to one of the twin networks and the 6 degree global pose information of the camera can be estimated.

We use the same image pre-processing approaches as previous methods \cite{kendall2015posenet}. We firstly resize the image to 256 pixels along the shorter side and normalize it with the mean and standard deviation computed from the ImageNet dataset. For the training phase, we randomly crop the image to $224 \times 224$ pixels. For the testing phase, images are cropped to  $224 \times 224$ pixels at the center of the image. Training images are shuffled before they are fed to the network.

The modified ResNet50 is initialized with pre-trained weights of ImageNet dataset. The GPRU component and the RPRU are initialized with the Xavier initialization \cite{glorot2010understanding}.
We choose the Adam optimizer to train the network with parameters $\beta_{1} = 0.9$ and $\beta_{2} = 0.999$. The weight decay is $10^{-5}$. We train the network with a learning rate of $10^{-5}$ and the batch-size is set to be 32. We initialize the $\hat{s}_{x}$ and $\hat{s}_{q}$ with  0 and -3.0 respectively in our experiments. We implement the network with PyTorch and train the network on an Ubuntu 16.04 TS system with a NVIDIA GTX 1080Ti GPU. 
Training is stopped until the network is converged.

\subsection{Results}
We compare the results of the proposed method with that of state-of-the-art deep learning-based methods such as PoseNet, Bayesian PoseNet, PoseNet2, Hourgrlass-net, LSTM-Net and RelNet on the \emph{7Scene} dataset, and with PoseNet, Bayesian PoseNet, PoseNet2 and LSTM-Net on the \emph{Cambridge Landmarks} dataset.
Similar to others, we report each scene's median error. We also compare the average median accuracy over all scenes in each dataset.
The comparative results are shown in Table \ref{tab:comparison_7Scenes} and Table \ref{tab:comparison_Cambridge}.
\begin{table*}[!ht]
	\caption{Comparison of median errors with other deep learning-based methods on the \emph{7Scene} dataset.}
	\resizebox{\textwidth}{!}{
		\begin{tabular}{lcccccccl}
			\hline
			Scene& PoseNet & Bayesian PoseNet& LSTM-Net& Vidloc & HourglassNet& PoseNet2 & Relnet & Ours \\
			\hline
			Chess & 0.32m,~$8.12^\circ$ & 0.37m,~$7.24^\circ$ &0.24m,~$5.77^\circ$ &0.18m,~N/A& 0.15m,~ $6.53^\circ$ & 0.13m,~$\textbf{4.48}^\circ$ &0.13m,~ $6.46^\circ$ & \textbf{0.099}m,~$5.19^\circ$ \\
			Fire & 0.47m,~$14.4^\circ$ & 0.43m,~$13.7^\circ$ &0.34m,~$11.9^\circ$ &0.26m,~N/A & 0.27m,~$\textbf{10.84}^\circ$& 0.27m,~$11.3^\circ$ &0.26m,~$12.72^\circ$ & \textbf{0.253}m,~$11.64^\circ$ \\
			Heads & 0.29m,~$12.0^\circ$ & 0.31m,~$12.0^\circ$ &0.21m,~$13.7^\circ$ &0.14m,~N/A& 0.19m,~$\textbf{11.63}^\circ$ & 0.17m,~$13.0^\circ$ &0.14m,~$12.34^\circ$ &\textbf{0.126}m,~$13.20^\circ$ \\
			Office & 0.48m,~$7.68^\circ$ & 0.48m,~$8.04^\circ$ &0.30m,~$8.08^\circ$ &0.26m,~N/A& 0.21m,~ $8.48^\circ$ & 0.19m,~$\textbf{5.55}^\circ$ &0.21m,~ $7.35^\circ$ &\textbf{0.161}m,~$7.71^\circ$ \\
			Pumpkin & 0.47m,~$8.42^\circ$ & 0.61m,~$7.08^\circ$ &0.33m,~$7.00^\circ$&0.36m,~N/A& 0.25m,~ $7.01^\circ$ & 0.26m,~$\textbf{4.75}^\circ$ &0.24m,~ $6.35^\circ$ &\textbf{0.163}m,~$6.61^\circ$ \\
			Redkitchen & 0.59m,~$8.64^\circ$ & 0.58m,~$7.54^\circ$ &0.37m,~$8.83^\circ$&0.31m,~N/A& 0.27m,~$10.15^\circ$ & 0.23m,~$\textbf{5.35}^\circ$ &0.24m,~ $8.03^\circ$ &\textbf{0.174}m,~$8.24^\circ$ \\
			Stairs & 0.47m,~$13.8^\circ$ & 0.48m,~$13.1^\circ$ &0.40m,~$13.7^\circ$&0.26m,~N/A& 0.29m,~$12.46^\circ$ & 0.35m,~$12.4^\circ$ &0.27m,~$\textbf{11.82}^\circ$ &\textbf{0.26}m,~$13.13^\circ$ \\
			\hline
			Average& 0.44m,~$10.4^\circ$ & 0.47m,~$9.81^\circ$ &0.31m,~ $9.85^\circ$&0.25m,~N/A& 0.23m,~ $9.53^\circ$ & 0.23m,~ $\textbf{8.12}^\circ$ &0.21m,~ $9.30^\circ$ &\textbf{0.177}m,~$9.39^\circ$ \\
			\hline
		\end{tabular}

	}
\label{tab:comparison_7Scenes}
\end{table*}
Table \ref{tab:comparison_7Scenes} shows the results for the \emph{7Scene} dataset. It is seen that compared with 7 state-of-the-art deep learning-based camera relocalization methods, the proposed method achieves the best performance on positional accuracy in all 7 scenes. 
Our method improves the average median positional accuracy by $16\%$ over the best reported result. It is interesting to note that our method has obtained even better result than PoseNet2, which utilizes 3D reference as additional constraints.

For orientational accuracy, we achieve the best result compared to methods based on direct regression. It is not surprising that the results are not as good as PoseNet2 and RelNet since PoseNet2 requires additional 3D models and RelNet triangulates the pose with all referencing images by estimating the relative poses instead of directly regressing results.

\begin{table*}[!ht]
	\caption{\label{tab:comparison_Cambridge}Comparison of median errors with other deep learning-based methods on the \emph{Cambridge Landmarks} dataset.}
	\centering
	\begin{tabular}{lcccccl}
		\hline
		Scene & PoseNet & Bayesian PoseNet & LSTM-Net& PoseNet2& Ours\\
		\hline
		KingsCollege&1.92m,~$5.40^\circ$ & 1.74m,~$4.06^\circ$ & 0.99m,~$3.68^\circ$ & 0.88m,~$\textbf{1.04}^\circ$& \textbf{0.865}m,~$1.96^\circ$\\
		OldHospital&2.31m,~$5.38^\circ$ & 2.57m,~$5.14^\circ$ & \textbf{1.51}m,~$4.29^\circ$ & 3.20m,~$3.29^\circ$& 1.617m,~$\textbf{2.42}^\circ$\\
		ShopFacade&1.46m,~$8.08^\circ$ & 1.25m,~$7.54^\circ$ & 1.18m,~$7.44^\circ$ & 0.88m,~$\textbf{3.78}^\circ$& \textbf{0.834}m,~$5.56^\circ$\\
		StMarysChurch&2.65m,~$8.46^\circ$ & 2.11m,~$8.38^\circ$ & \textbf{1.52}m,~$6.68^\circ$ & 1.57m,~$3.32^\circ$& 1.650m,~$\textbf{2.98}^\circ$\\
		\hline
		Average&2.08m,~$6.83^\circ$ & 1.92m,~$6.28^\circ$ & 1.30m,~$5.52^\circ$& 1.62m,~$\textbf{2.86}^\circ$ & \textbf{1.24}m,~$3.23^\circ$ \\
		\hline
	\end{tabular}
	
\end{table*}

Table \ref{tab:comparison_Cambridge} shows the results for the \emph{Cambridge Landmarks} dataset. It can be seen that our method obtains the best positional accuracy on the KingsCollege and the ShopFacade scenes, reaching accuracies of 0.865m and 0.834m respectively.
We improve the state-of-the-art orientational accuracy of the OldHospital and the StMarysChurch scenes from $3.29^\circ$ and $3.32^\circ$  to $2.42^\circ$ and $ 2.98^\circ$, achieving 26\% and 10\% improvement respectively.
The average positional accuracy over all scenes is improved from $1.30m$ to $1.24m$. The average orientational accuracy over all scenes is only a little worse than that of PoseNet2, which is trained with 3D model constraints.

It is interesting to note that of all the methods presented in the two tables, some did  better in positional accuracy and some did better in orientational accuracy, none of them seems to comprehensively beat the others in both measures. Our method achieves the best average positional accuracy amongst all methods in both datasets. For orientational accuracy, our method achieves competent results, which is only slightly worse than the best method (PoseNet2) but better or at least as good as the other methods.

\subsection{Discussion}
In this section, we perform analysis on the influence of various loss function components and the reference image selection strategy. The experiments are also done on the \emph{7Scene} and \emph{Cambridge Landmarks}.
\subsubsection{Loss Analysis}
We perform ablation analysis on the loss function. Recall from equation \eqref{init_overall_loss}, the overall loss function is $L = L_{G}+L_{C}+L_{R}+L_{MD}$, consisting of the the global loss $L_{G}$, the relative pose loss $L_{C}$, the relative pose regression loss $L_{R}$, and the adaptive metric distance loss $L_{MD}$. In order to assess the role these loss components play, we formulate  4 loss functions based on the following combinations:

\begin{enumerate}[(1)]
	\item G: GlobalLoss;
	\item G+C: GlobalLoss $+$ RelLoss;
	\item G+C+R: GlobalLoss $+$ RelLoss $+$ RelRLoss;
	\item Ours: GlobalLoss $+$ RelLoss $+$ RelRLoss $+$ MDLoss.
\end{enumerate}

\begin{table*}[!t]
	\caption{\label{tab:loss analysis:7Scene}Comparison of different loss combinations with median error on \emph{7Scene} dataset.}
	\centering
	\begin{tabular}{lcccccc}
		\hline
		Scene& G & G+C& G+C+R& Ours\\
		\hline
		Chess & 0.135m,~~~$7.62^\circ$ & 0.118m,~~~$\textbf{5.10}^\circ$ &0.116m,~~~$6.50^\circ$  &\textbf{0.099}m,~~~ $5.19^\circ$ \\
		Fire & 0.285m,~~$13.13^\circ$ & 0.258m,~$12.93^\circ$ &0.258m,~$12.48^\circ$ &\textbf{0.253}m, ~~$\textbf{11.64}^\circ$ \\
		Heads &0.185m,~~$14.01^\circ$ &0.140m,~$14.77^\circ$ &0.144m,~$13.82^\circ$ &\textbf{0.126}m, ~~$\textbf{13.20}^\circ$ \\
		Office & 0.180m,~~~$8.18^\circ$ & 0.173m,~~~$\textbf{7.65}^\circ$ &0.175m,~~~$8.19^\circ$ & \textbf{0.161}m,~~~ $7.71^\circ$ \\
		Pumpkin & 0.215m,~~~$7.77^\circ$ & 0.226m,~~~$7.87^\circ$ &0.214m,~~~$6.80^\circ$  &\textbf{0.163}m,~~~ $\textbf{6.61}^\circ$ \\
		Redkitchen & 0.266m,~~~$\textbf{8.21}^\circ$ & 0.253m,~~~$9.20^\circ$ &0.201m,~~~$8.24^\circ$  &\textbf{0.174}m,~~~$8.24^\circ$ \\
		Stairs & 0.345m,~$13.51^\circ$ & 0.324m,~$\textbf{12.07}^\circ$ &0.279m,~$13.18^\circ$  &\textbf{0.260}m,~~$13.13^\circ$ \\
		\hline
		Average& 0.230m,~$10.34^\circ$ & 0.213m,~~~$9.94^\circ$ &0.198m,~~~$9.89^\circ$ &\textbf{0.177}m,~~~ $\textbf{9.39}^\circ$ \\
		\hline
	\end{tabular}
\end{table*}

%

\begin{figure}[htbp]
	
	\centering
	\includegraphics[width=80mm]{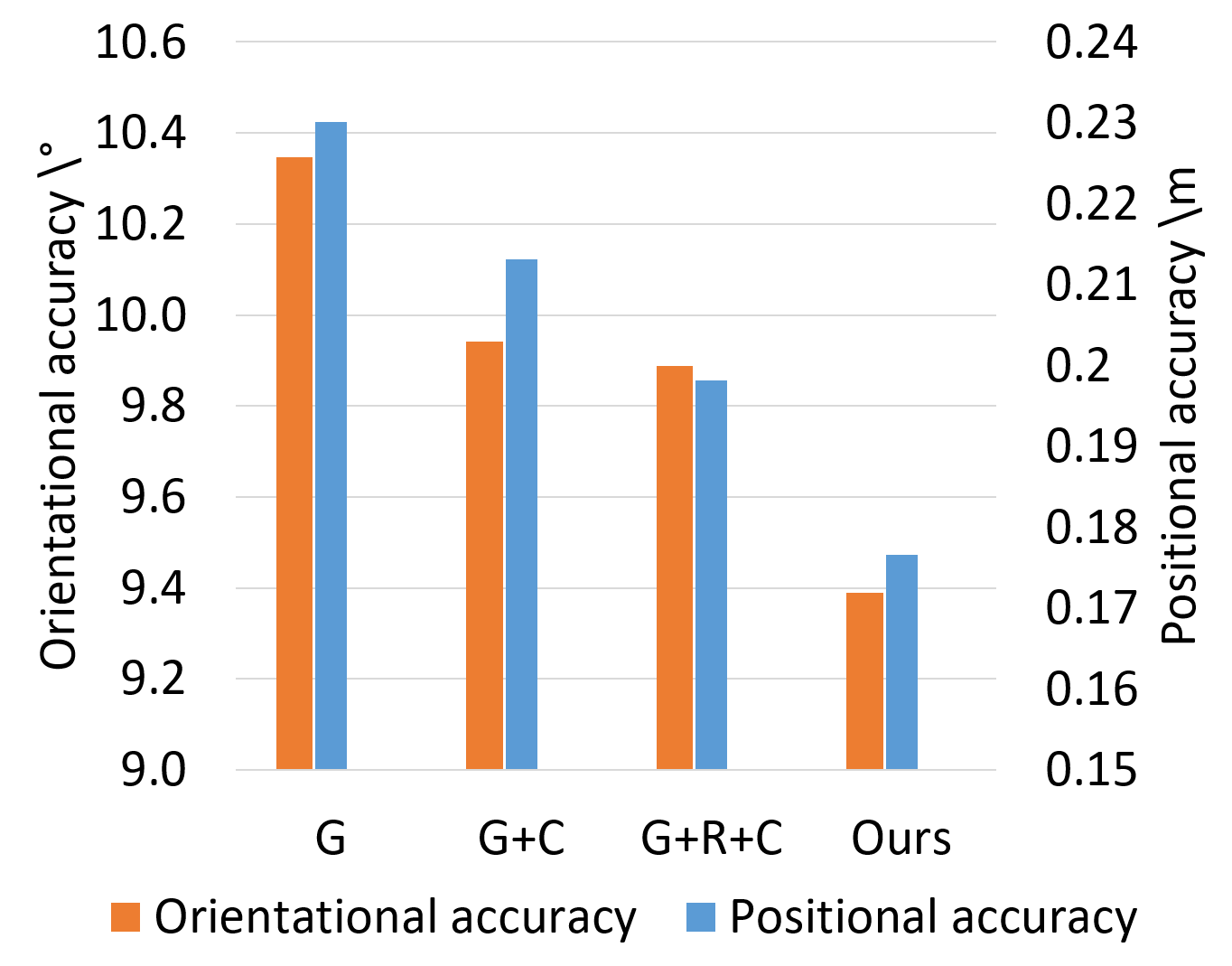}
	\caption{The loss analysis over the average errors on \emph{7Scene} dataset.}
	\label{fig:Loss_Pos:7Scene}
\end{figure}

\begin{figure}[htbp]
	
	\centering
	\includegraphics[width=80mm,height=45mm]{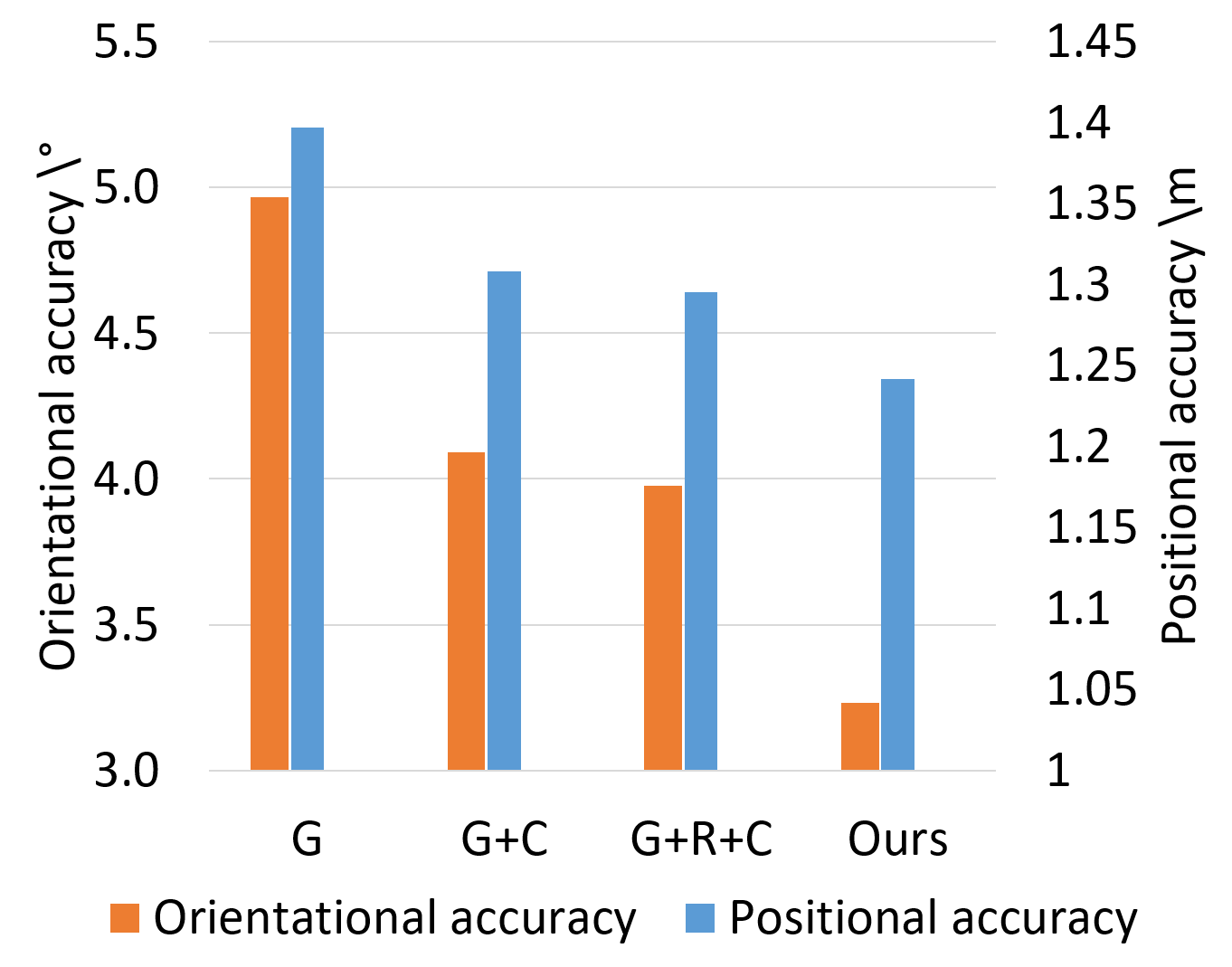}
	\caption{The loss analysis over the average errors on \emph{Cambridge Landmarks} dataset.}
	\label{fig:Loss_Pos:Cam}
\end{figure}

We train the proposed network by the 4 aforementioned loss functions separately. The results are shown in Table \ref{tab:loss analysis:7Scene} for the  \emph{7Scene} dataset and in Table \ref{tab:loss analysis:Cam} for \emph{Cambridge landmarks}. It is seen  that as more loss terms are added to the loss function, both positional error and orientational error decrease for all scenes of the \emph{7Scene} dataset and the \emph{Cambridge Landmarks} dataset.
The average positional error and orientational error for the \emph{7Scene} dataset and \emph{Cambridge Landmarks} dataset are shown in Figure \ref{fig:Loss_Pos:7Scene} and in Figure \ref{fig:Loss_Pos:Cam} respectively. We can see that average position and orientation errors show a decreasing trend by adding more constraints. This demonstrates the usefulness of each loss component combinations.

\begin{table*}[!ht]
	\caption{\label{tab:loss analysis:Cam}Comparison of different loss combinations with median error on \emph{Cambridge Landmarks} dataset.}
	\centering
	\begin{tabular}{lcccccccccccl}
		\hline
		Scene& G & G+C& G+C+R& Ours\\
		\hline
		KingsCollege & 1.07m,~~~$4.22^\circ$ & 0.932m,~~~$2.69^\circ$ &0.97m,~~~$2.14^\circ$  &\textbf{0.865}m,~~~$\textbf{1.96}^\circ$ \\
		OldHospital & 1.76m,~~~$4.97^\circ$ & 1.650m,~~~$3.38^\circ$ &1.67m,~~~$3.01^\circ$  &\textbf{1.617}m,~~~$\textbf{2.42}^\circ$ \\
		ShopFacade & 1.00m,~~~$6.65^\circ$ & 0.930m,~~~$6.23^\circ$&0.858m,~~~$5.92^\circ$ &\textbf{0.834}m,~~~$\textbf{5.56}^\circ$ \\
		StMarysChurch & 1.76m,~~~$4.03^\circ$ & 1.720m,~~~$4.06^\circ$ &1.684m,~~~$4.83^\circ$ &\textbf{1.615}m,~~~$\textbf{2.98}^\circ$ \\
		\hline
		Average& 1.396m,~~~$4.97^\circ$ & 1.308m,~~~$4.09^\circ$ &1.296m,~~~$3.98^\circ$ &\textbf{1.242}m,~~~$\textbf{3.23}^\circ$ \\
		\hline
	\end{tabular}
\end{table*}

%

\subsubsection{Comparison of Relative Geometry Losses }
We have designed three relative geometry-based losses. In order to evaluate their performance separately for pose prediction, we formulate new losses by combining each of them with the global pose loss. We also use global pose loss and our comprehensive loss as baselines. The details of the loss combinations are listed as follows:

\begin{enumerate}[(1)]
	\item G  : GlobalLoss;
	\item G+M: GlobalLoss $+$ MDLoss;
	\item G+C: GlobalLoss $+$ RelLoss;
	\item G+R: GlobalLoss $+$ RelRLoss;
	\item Ours: GlobalLoss $+$ RelLoss $+$ RelRLoss $+$ MDLoss.
\end{enumerate}

\begin{figure}[htbp]
	
	\centering
	\includegraphics[width=80mm,height=45mm]{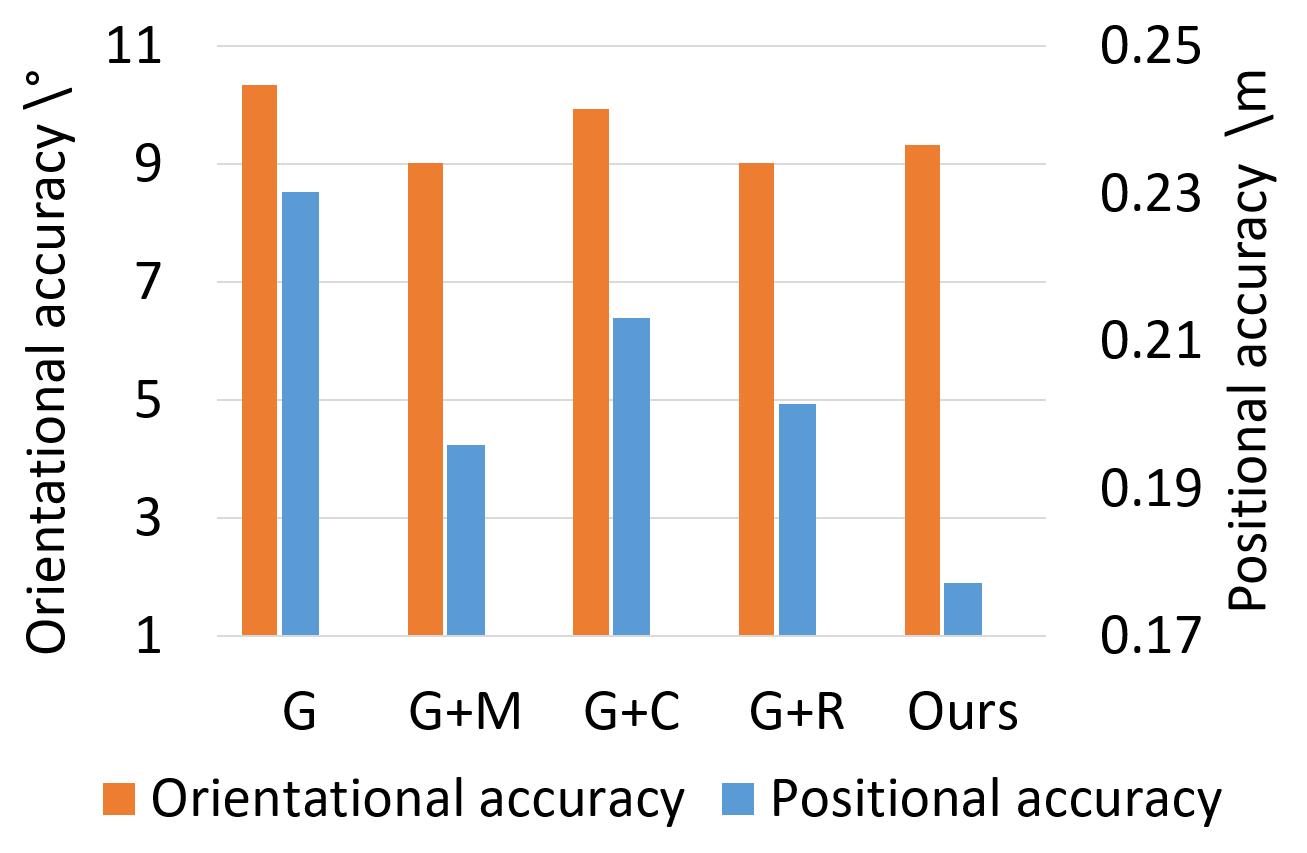}
	\caption{The relative loss analysis over the average errors on \emph{7Scene} dataset.}
	\label{fig:Rel_Pos:7Scene}
\end{figure}

\begin{figure}[htbp]
	
	\centering
	\includegraphics[width=80mm,height=45mm]{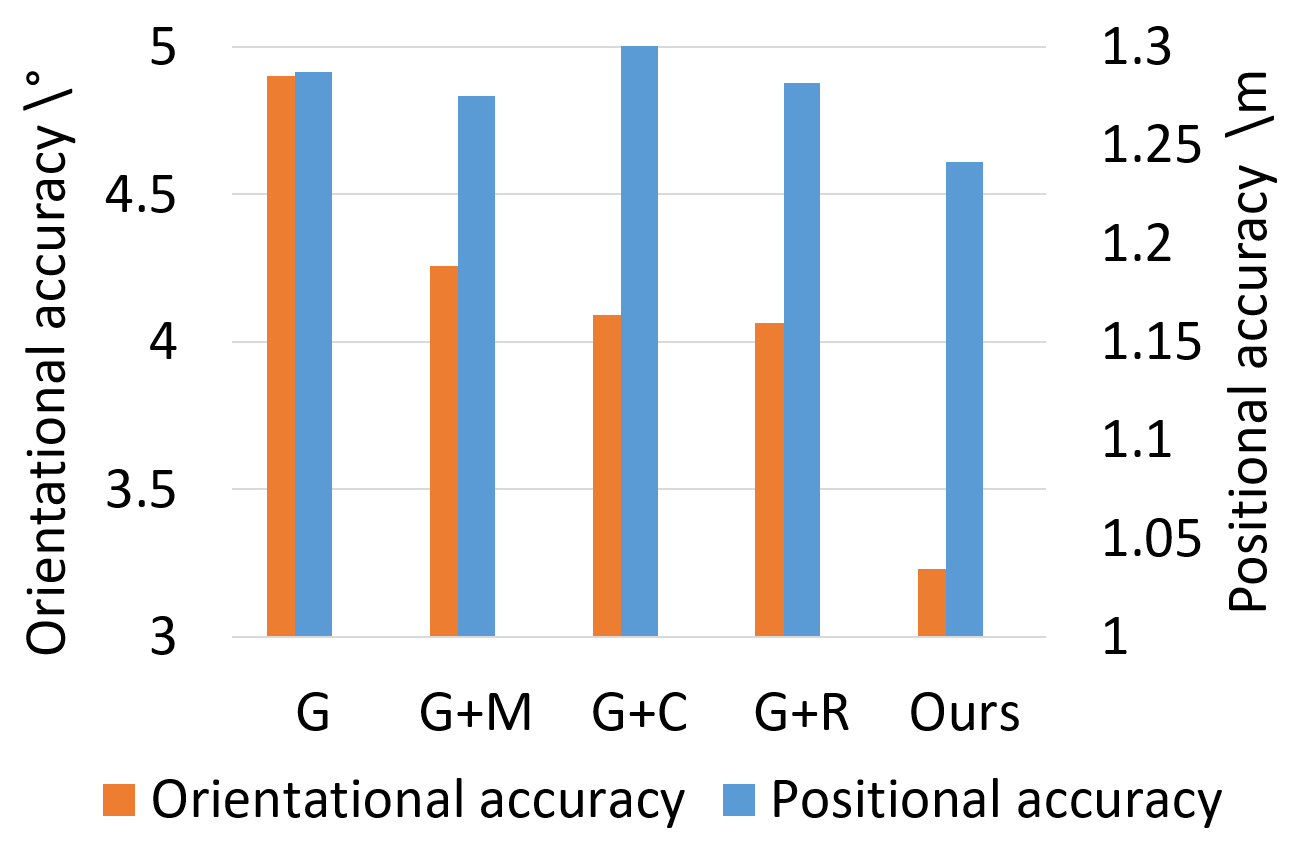}
	\caption{The relative loss analysis over the average errors on \emph{Cambridge Landmarks} dataset.}
	\label{fig:Rel_Pos:Cam}
\end{figure}

For each loss function, we repeat experiments using the same training setup in previous experiments. The results on \emph{7Scene} and on \emph{Cambridge Landmarks} are shown in Table \ref{tab:rel_loss analysis:7Scene} and in Table \ref{tab:rel_loss analysis:Cam} respectively.
The average localization errors of the two datasets are shown in Figure \ref{fig:Rel_Pos:7Scene} and Figure \ref{fig:Rel_Pos:Cam}.
\begin{table*}[!t]
	\caption{\label{tab:rel_loss analysis:7Scene}Evaluation of each relative loss function with median error on \emph{7Scene} dataset.}
	\centering
	\begin{tabular}{lcccccccccccl}
		\hline
		Scene& G & G+M & G+C& G+R& Ours\\
		\hline
		Chess & 0.135m,~~~$7.62^\circ$ & 0.116m,~~~$\textbf{4.82}^\circ$ & 0.118m,~~~$5.10^\circ$ &0.117m,~~~$5.05^\circ$  & \textbf{0.099}m,~~~ $5.19^\circ$ \\
		Fire & 0.285m,~$13.13^\circ$ & 0.271m,~$11.91^\circ$ & 0.258m,~$12.93^\circ$ &0.262m,~$12.64^\circ$  & \textbf{0.253}m,~~~$\textbf{11.64}^\circ$ \\
		Heads & 0.185m,~$14.01^\circ$ & 0.128m,~$13.37^\circ$ &0.140m,~$14.77^\circ$ &0.147m,~ $13.21^\circ$ & \textbf{0.126}m,~~~$\textbf{13.20}^\circ$ \\
		Office & 0.180m,~~~$8.18^\circ$ &0.177m,~~~$7.17^\circ$ & 0.173m,~~~$7.65^\circ$ &0.189m,~~~$\textbf{7.13}^\circ$ &\textbf{0.161}m,~~~ $7.71^\circ$ \\
		Pumpkin & 0.215m,~~~$7.77^\circ$ &0.198m,~~~$6.26^\circ$ & 0.226m,~~~$7.87^\circ$ &0.196m,~~~$\textbf{5.82}^\circ$ &\textbf{0.163}m,~~~ $6.61^\circ$ \\
		Redkitchen & 0.266m,~~~$8.21^\circ$ & 0.217m,~~~$\textbf{7.55}^\circ$ & 0.253m,~~~$9.20^\circ$ &0.218m,~~~$7.79^\circ$  &\textbf{0.174}m,~~~ $8.24^\circ$ \\
		Stairs & 0.345m,~$13.51^\circ$ & 0.265m,~$11.98^\circ$ & 0.324m,~$12.07^\circ$ &0.281m,~$\textbf{11.49}^\circ$  & \textbf{0.260}m,~~$13.13^\circ$ \\
		\hline
		Average& 0.230m,~$10.34^\circ$ & 0.196m,~$\textbf{9.01}^\circ$ & 0.213m,~~~$9.94^\circ$ &0.201m,~~~$9.02^\circ$ &\textbf{0.177}m,~~~ $9.39^\circ$ \\
		\hline
	\end{tabular}
\end{table*}

\begin{table*}[!ht]
	\caption{\label{tab:rel_loss analysis:Cam}Evaluation of each relative loss function with median error on \emph{Cambridge Landmarks} dataset.}
	\centering
	\begin{tabular}{lcccccccccccl}
		\hline
		Scene& G &G+M & G+C& G+R& Ours\\
		\hline
		KingsCollege & 1.07m,~~~$4.22^\circ$ & 0.960m,~~~$2.79^\circ$ & 0.932m,~~~$2.69^\circ$ &0.980m,~~~$2.31^\circ$  &\textbf{0.865}m,~~~$\textbf{1.96}^\circ$ \\
		OldHospital & 1.76m,~~~$4.97^\circ$ & 1.650m,~~~$3.31^\circ$ & 1.650m,~~~$3.38^\circ$ &\textbf{1.615}m,~~~$3.77^\circ$  &1.617m,~~~$\textbf{2.42}^\circ$ \\
		ShopFacade & 1.00m,~~~$6.65^\circ$ & 0.876m,~~~$\textbf{5.11}^\circ$ &0.930m,~~~$6.23^\circ$&0.868m,~~~$5.19^\circ$ &\textbf{0.834}m,~~~$5.56^\circ$ \\
		StMarysChurch & 1.76m,~~~$4.03^\circ$ & 1.617m,~~~$5.83^\circ$ & 1.720m,~~~$4.06^\circ$ &1.664m,~~~$4.68^\circ$ &\textbf{1.615}m,~~~ $\textbf{2.98}^\circ$ \\
		\hline
		Average& 1.396m,~~~$4.97^\circ$ &1.275m,~~~$4.257^\circ$ & 1.308m,~~~$4.09^\circ$ &1.282m,~~~$4.06^\circ$ &\textbf{1.242}m,~~~ $\textbf{3.23}^\circ$ \\
		\hline
	\end{tabular}
\end{table*}

As shown in the two Figures, relative geometry-related losses (G+M, G+C, G+R) achieve better accuracy than global pose alone in every scene of the two datasets. This further demonstrates their effectiveness on global pose prediction. It  can also be seen that G+M obtains a larger average accuracy increase compared with the other two. In addition, G+C acquires the smallest accuracy improvement on both datasets, lower than G+R. This implies that relative geometry constraints work better in feature space than in the pose space since RelRLoss and MDLoss are in the feature space while RelLoss is in the pose space.
It should also be noted that the results of our proposed loss (G+C+R+M) outperforms all the other single relative geometry-related losses, which further demonstrate the effectiveness of our comprehensive loss function.

\subsubsection{Comparison of Metric Losses}

To further evaluate the proposed adaptive metric distance loss, we conduct experiments to compare it with conventional siamese loss \cite{hadsell2006dimensionality} and triplet loss \cite{Schroff_2015_CVPR}, since the two losses can also help 
make visually similar image distinctive as the proposed metric distance loss does. The siamese loss is shown in equation \eqref{siamese} and the triplet loss is shown in equation \eqref{triplet}. The major difference is that the conventional metric losses set the margin to be a fixed value while our loss is a function of the relative pose of two images.

\begin{equation}\label{siamese}
L_{Siamese} =  \frac{1}{2N}\sum_{n=1}^{N} \{(1-y)d^2+y\{max(m - d,0)\}^2 \},
\end{equation}

\begin{equation}\label{triplet}
L_{Triplet} =  \sum_{n=1}^{N}[\left\| f(x_{i}^{a})-f(x_{i}^{p})\right\|_{2}^{2}-\left\| f(x_{i}^{a})-f(x_{i}^{n})\right\|_{2}^{2}+m]_{+}
\end{equation}
where $[x]_{+}$ represents $ max(x,0)$ as hinge loss, $N$ is the number of training samples, $m$ is the margin, $d$ is the feature distance of the paired image, $y$ always equals 1, since the two images are not from the same location.  $f(x_{i}^{a})$, $f(x_{i}^{p})$ and $f(x_{i}^{n})$ are the feature vectors of the $i$th training image, its reference images, and the image after the reference image, respectively. In the siamese loss of equation \eqref{siamese}, it explicitly forces the features of the two images to be different because they are from two different locations. In the triplet loss \eqref{triplet}, it explicitly enforces that the difference between the $i$th image and its reference should be smaller than the difference between it and the image after the reference image.
In the experiments, we simply replace the MDLoss with the siamese loss and the triplet loss respectively and repeat the experiment. The margin parameter $m$ of the siamese loss and the triplet loss is empirically set to be 0.001, which gives the best accuracy. Three comparative losses are listed as below.

\begin{enumerate}[(1)]
	\item LossSiamese: GlobalLoss $+$ RelLoss $+$ RelRLoss $+$ SiameseLoss;
	\item LossTriplet: GlobalLoss $+$ RelLoss $+$ RelRLoss $+$ TripletLoss;
	\item Ours:        GlobalLoss $+$ RelLoss $+$ RelRLoss $+$ MDLoss.
\end{enumerate}

\begin{figure}[htbp]
	
	\centering
	\includegraphics[width=80mm]{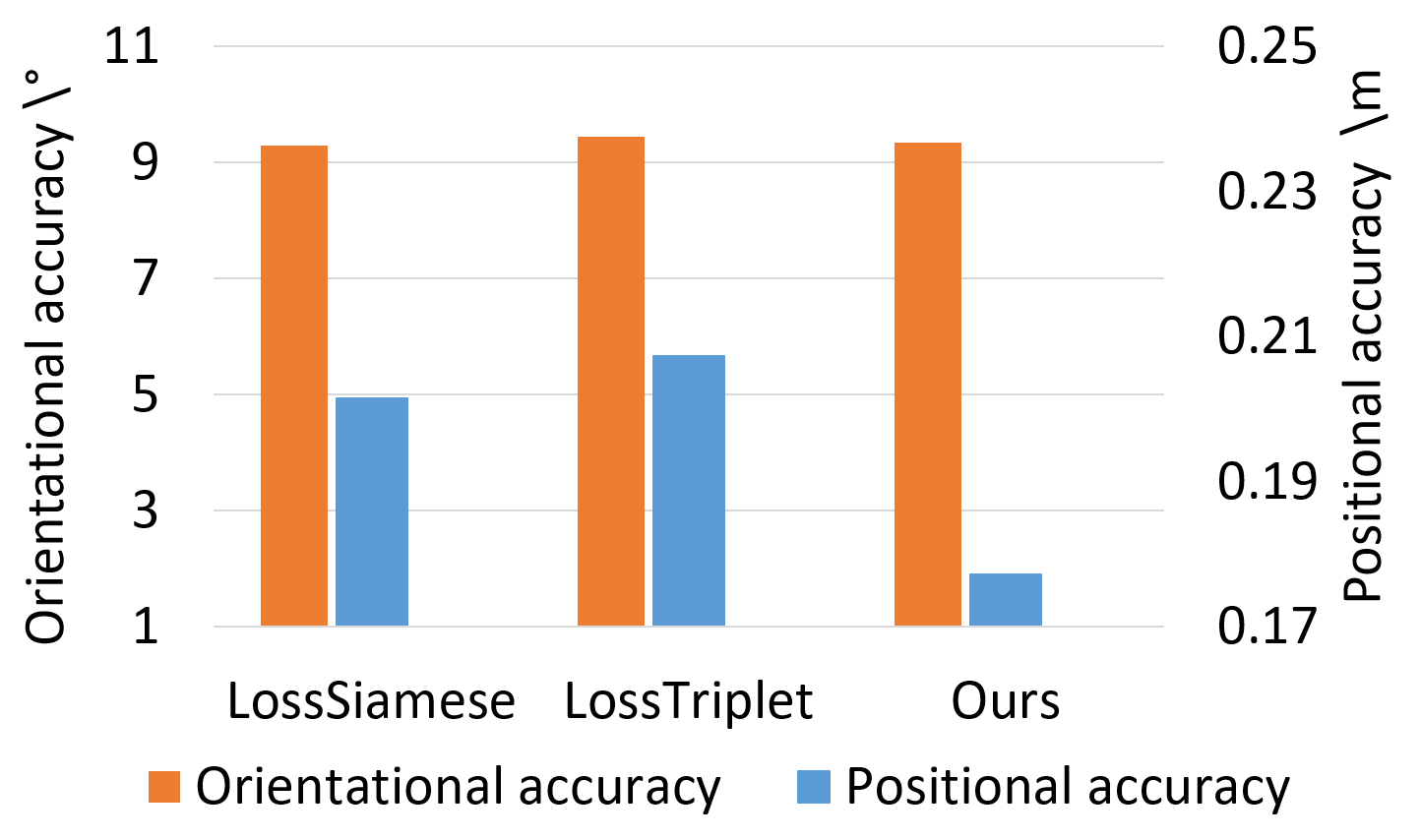}
	\caption{The metric loss analysis over the average errors on \emph{7Scene} dataset.}
	\label{fig:Metr_Pos:7Scene}
\end{figure}

\begin{figure}[htbp]
	
	\centering
	\includegraphics[width=80mm]{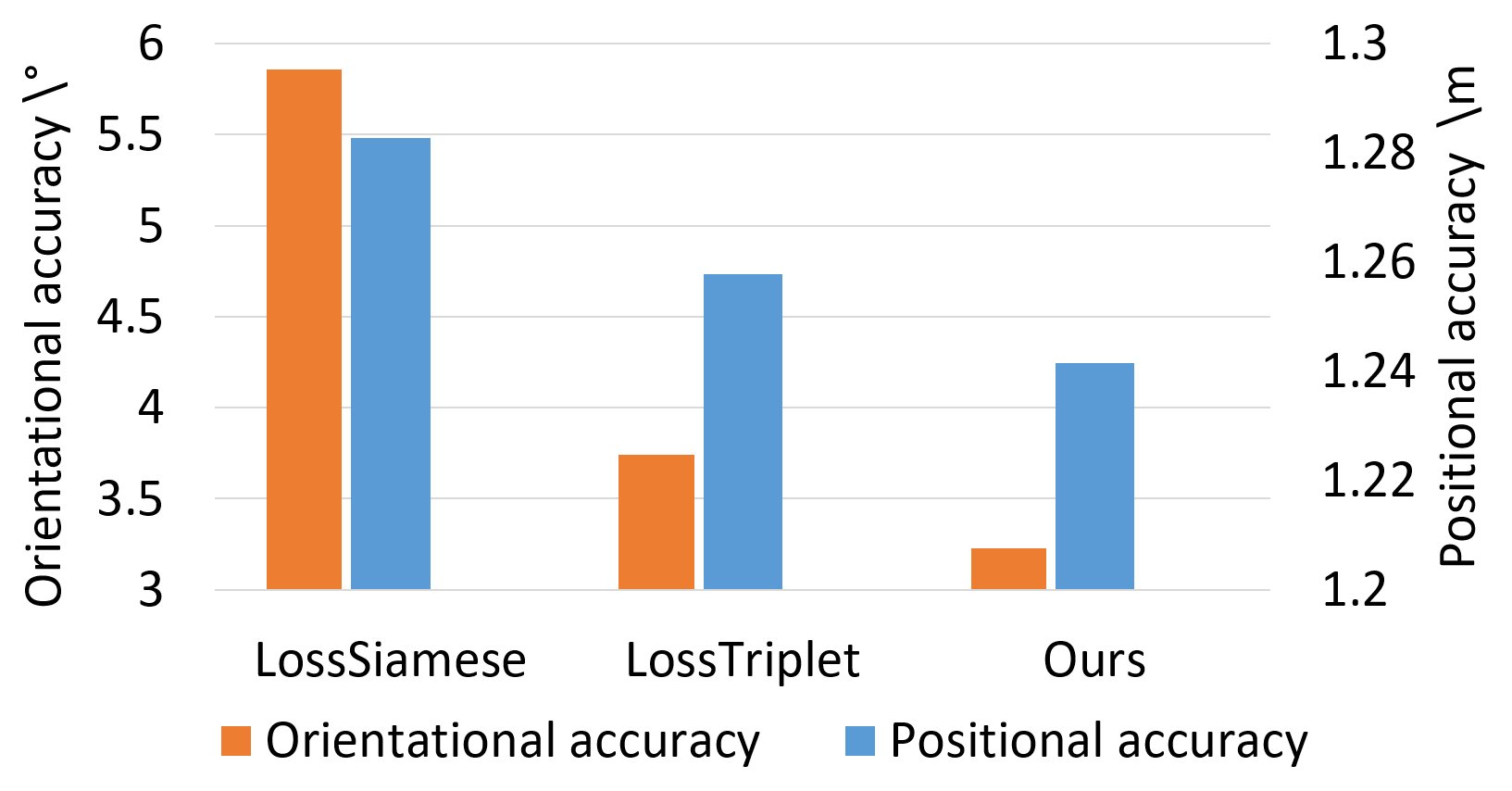}
	\caption{The metric loss analysis over the average errors on \emph{Cambridge Landmarks} dataset.}
	\label{fig:Metr_Pos:Cam}
\end{figure}

We repeat the experiments on the two datasets using the above losses and the results are shown in Table \ref{tab:metric_loss analysis:7Scene} for \emph{7Scene} and Table \ref{tab:metric_loss analysis:Cam} for \emph{Cambridge Landmarks}. The average localization errors of the two datasets are shown in Figure \ref{fig:Metr_Pos:7Scene} and Figure \ref{fig:Metr_Pos:Cam}.

\begin{table}[!t]
	\caption{\label{tab:metric_loss analysis:7Scene}Comparison between metric loss functions and adaptive metric distance loss with median error on \emph{7Scene} dataset.}
	\centering
	\begin{tabular}{lcccccccccccl}
		\hline
		Scene&  LossSiamese & LossTriplet&  Ours\\
		\hline
		Chess      & 0.127m,~~~$\textbf{5.17}^\circ$  & 0.139m,~~~$5.42^\circ$    &\textbf{0.099}m,~~~$5.19^\circ$ \\
		Fire       & 0.273m,~$13.57^\circ$   & 0.276m,~$12.94^\circ$ &\textbf{0.253}m,~~$\textbf{11.64}^\circ$ \\
		Heads      & 0.128m,~$13.45^\circ$  &\textbf{0.125}m,~$14.76^\circ$ &0.126m,~~$\textbf{13.20}^\circ$ \\
		Office     & 0.188m,~~~$7.77^\circ$ & 0.192m,~~~$7.60^\circ$    &\textbf{0.161}m,~~~ $\textbf{7.71}^\circ$ \\
		Pumpkin    & 0.198m,~~~$6.13^\circ$  & 0.216m,~~~$\textbf{6.03}^\circ$    &\textbf{0.163}m,~~~ $6.61^\circ$ \\
		Redkitchen & 0.219m,~~~$8.32^\circ$  & 0.224m,~~~$8.30^\circ$    &\textbf{0.174}m,~~~ $\textbf{8.24}^\circ$ \\
		Stairs     & 0.277m,~~$\textbf{10.59}^\circ$  & 0.279m,~$11.07^\circ$ &\textbf{0.260}m,~~$13.13^\circ$ \\
		\hline
		Average    & 0.201m,~~$\textbf{9.29}^\circ$      & 0.207m,~~~$9.44^\circ$    &\textbf{0.177}m,~~~ $9.39^\circ$ \\
		\hline
	\end{tabular}
\end{table}

\begin{table}[!ht]
	
	\caption{\label{tab:metric_loss analysis:Cam}Comparison between metric loss functions and adaptive metric distance loss with median error on \emph{Cambridge Landmarks} dataset.}
	\centering
	\begin{tabular}{lcccccccccccl}
		\hline
		Scene& LossSiamese & LossTriplet&  Ours\\
		\hline
		KingsCollege & 0.867m,~~~$4.87^\circ$ & \textbf{0.839}m,~~~$2.03^\circ$  &0.865m,~~~$\textbf{1.96}^\circ$ \\
		OldHospital & 1.675m,~~~$5.73^\circ$ & 1.683m,~~~$3.82^\circ$ &\textbf{1.617}m,~~~$\textbf{2.42}^\circ$ \\
		ShopFacade & 0.861m,~~~$5.76^\circ$ &0.847m,~~~$\textbf{5.01}^\circ$ &\textbf{0.834}m,~~~$5.56^\circ$ \\
		StMarysChurch & 1.728m,~~~$7.06^\circ$ & 1.650m,~~~$4.10^\circ$  &\textbf{1.615}m,~~~$\textbf{2.98}^\circ$ \\
		\hline
		Average& 1.282m,~~~$5.86^\circ$ & 1.258m,~~~$3.74^\circ$  &\textbf{1.242}m,~~~$\textbf{3.23}^\circ$ \\
		\hline
	\end{tabular}
\end{table}

It can be seen that our method achieves the best average position accuracy on \emph{7Scene} dataset, and both average position accuracy and average orientation accuracy on the \emph{Cambridge Landmarks} dataset. The LossSiamese acquires the best orientational accuracy on the \emph{7Scene} dataset. LossTriplet performs badly on the \emph{7Scene} dataset but obtains better performance than LossSiamese on the \emph{Cambridge Landmarks} dataset. Although the LossSiamese achieves the best orientational accuracy, our method obtains more best performances on each scene of the two datasets shown in Table \ref{tab:metric_loss analysis:7Scene} and in Table \ref{tab:metric_loss analysis:Cam}.
The results show that our adaptive metric distance loss outperforms the conventional siamese loss and the triplet loss.

\subsubsection{Reference Image Analysis}

In this section, we evaluate two strategies of choosing the reference image. One obvious strategy is to pair every two different images, but it will result in exponential increase of the training time and high information redundancy. 
To make the training phase efficient, we generate only one reference image for each image. Specifically, reference images are selected in two ways: 1) select the next image in the same image sequence as the reference image; 2) randomly select a different image of the dataset that is not a reference image of any other images. It should be noted that the next image is visually similar to the current image. Randomly chosen reference image has no such property.
To evaluate the effectiveness of the two reference image selection strategies on the adaptive metric loss (MDLoss), we train the proposed network with the comprehensive loss function. 
In addition, we use the result of the networks trained without MDLoss (G+R+C) as baseline to compare the results.
%
%

\begin{table}[htbp]
	\caption{\label{tab:refence performance:7scene}Comparison of median errors of two reference image selection strategies on the \emph{7Scene} dataset.}
	\centering
	
	\begin{tabular}{lcccccccccccl}
		\hline
		Scene& G+R+C& Random & Next\\
		\hline
		Chess &0.116m,~ $6.50^\circ$ & 0.109m,~ $5.46^\circ$  &\textbf{0.099}m,~ $\textbf{5.19}^\circ$ \\
		Fire &0.258m,~$12.48^\circ$ & 0.265m,~$12.54^\circ$ &\textbf{0.253}m,~$\textbf{11.64}^\circ$ \\
		Heads &0.144m,~$13.82^\circ$ & 0.138m,~$13.72^\circ$ &\textbf{0.126}m,~$\textbf{13.20}^\circ$\\
		Office &0.175m,~ $8.19^\circ$ & 0.172m,~ $8.17^\circ$ & \textbf{0.161}m,~ $\textbf{7.71}^\circ$ \\
		Pumpkin &0.214m,~ $6.80^\circ$ & 0.207m,~ $\textbf{6.33}^\circ$ &\textbf{0.163}m,~ $6.61^\circ$ \\
		Redkitchen &0.201m,~ $8.24^\circ$ & 0.202m,~ $8.89^\circ$ &\textbf{0.174}m,~ $\textbf{8.24}^\circ$ \\
		Stairs &0.279m,~$13.18^\circ$ & 0.287m,~$\textbf{11.89}^\circ$ & \textbf{0.260}m,~$13.13^\circ$\\
		\hline
		Average &0.198m,~ $9.89^\circ$ & 0.197m,~ $9.57^\circ$ & \textbf{0.177}m,~$\textbf{9.39}^\circ$ \\
		\hline
	\end{tabular}
	
\end{table}

\begin{figure}[htbp]
	
	\centering
	\includegraphics[width=80mm]{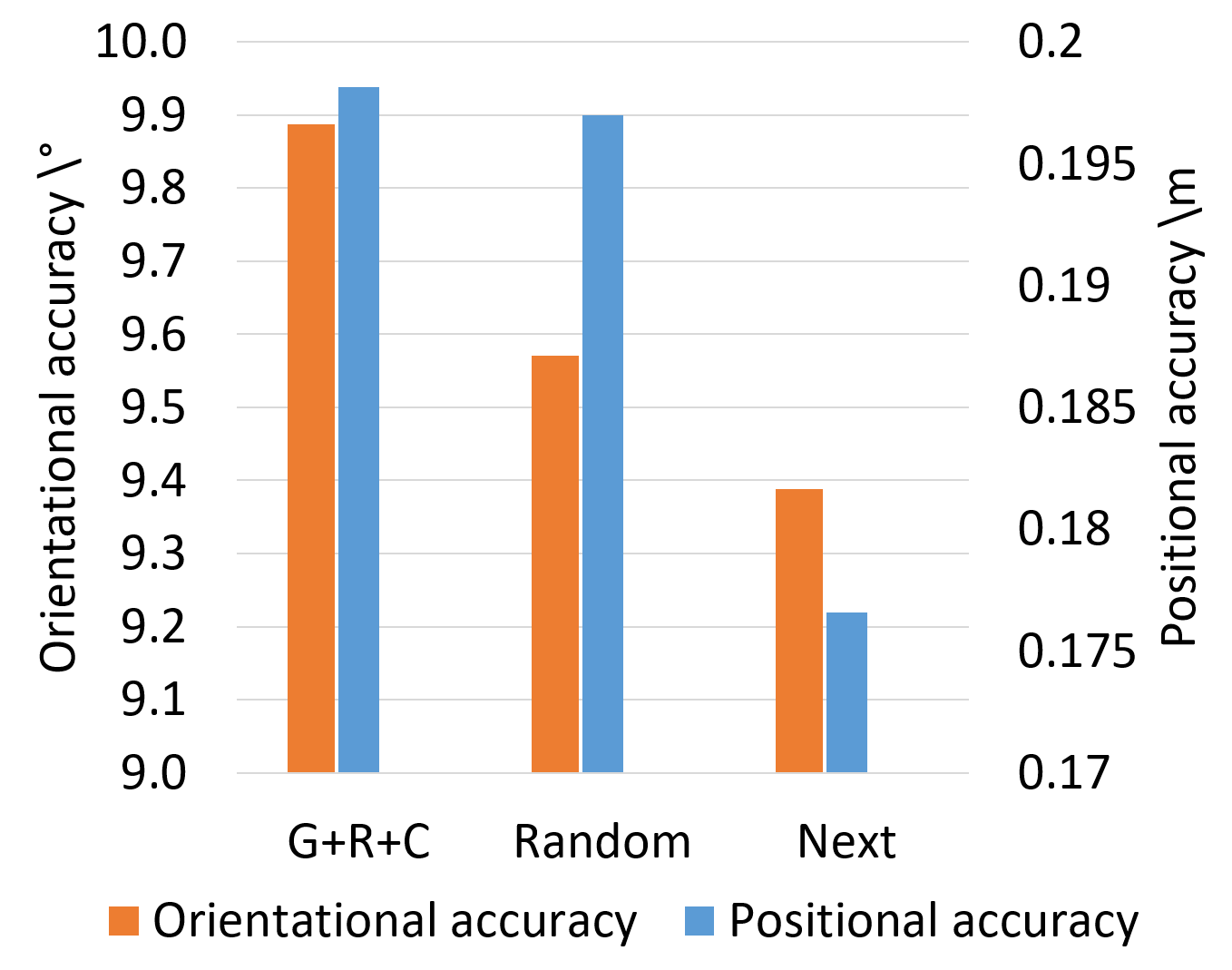}
	\caption{The average median errors of two reference image selection strategies on \emph{7Scene} dataset.}
	\label{fig:Ref_Pos:7scene}
\end{figure}
\begin{figure}[htbp]
	\centering
	\includegraphics[width=80mm]{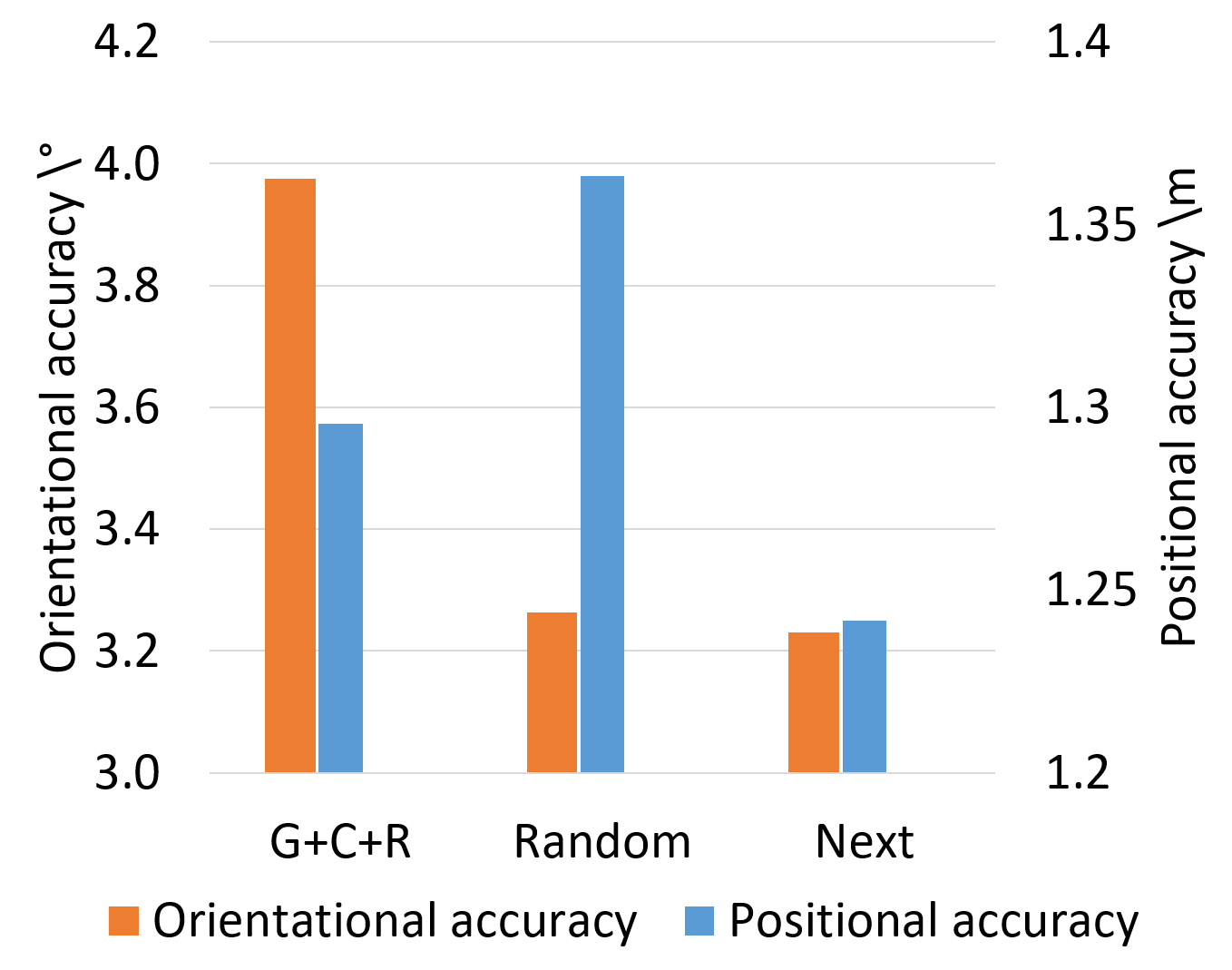}
	\caption{The average median errors of two reference image chosen strategies on \emph{Cambridge Landmarks}.}
	\label{fig:Ref_Pos:Cam}
\end{figure}

Comparative median error results for the different reference selection strategies for the \emph{7Scene}  and \emph{Cambridge Landmarks} are shown in Table \ref{tab:refence performance:7scene} and  Table \ref{tab:refence performance:Cam}. The average positional and orientational errors are shown in Figure \ref{fig:Ref_Pos:7scene} and in Figure \ref{fig:Ref_Pos:Cam}. It can be seen that strategy of choosing the next image as reference image obtains higher image similarity score than that of randomly choosing in two datasets since it achieves lower feature distance.

From Table \ref{tab:refence performance:7scene}, it is seen that compared to the random reference selection strategy, taking the next image as reference image increases the average positional accuracy from 0.197m to 0.177m and the average orientational accuracy from $9.57^\circ$ to $9.39^\circ$. It is also seen that for both reference image selection strategies, the inclusion of MDLoss improves performance. One probable explanation is that MDLoss makes the network learn to keep similar images of different poses apart in the feature space. 


%
\begin{table}[htbp]
	\caption{\label{tab:refence performance:Cam}Comparison of median errors of two reference image selection strategies on \emph{Cambridge Landmarks} dataset.}
	\centering
	
	\begin{tabular}{lccccccl}
		\hline
		Scene& G+C+R& Random & Next\\
		\hline
		KingsCollege    &0.970m,~$2.14^\circ$ & 1.120m,~$2.09^\circ$  &\textbf{0.865}m,~$\textbf{1.96}^\circ$ \\
		OldHospital     &1.670m,~$3.01^\circ$ & 1.618m,~$2.80^\circ$  &\textbf{1.617}m,~$\textbf{2.42}^\circ$ \\
		ShopFacade      &0.858m,~$5.92^\circ$ & 1.000m,~$\textbf{4.91}^\circ$ &\textbf{0.834}m,~$5.56^\circ$\\
		StMarysChurch   &1.684m,~$4.83^\circ$ & 1.714m,~$3.26^\circ$ &\textbf{1.650}m,~$\textbf{2.98}^\circ$ \\
		\hline
		Average         &1.296m,~$3.98^\circ$ & 1.363m,~$3.26^\circ$ &\textbf{1.242}m,~$\textbf{3.23}^\circ$ \\
		\hline
	\end{tabular}
	
\end{table}

\begin{figure}[htbp]
	\centering
	\includegraphics[width=80mm]{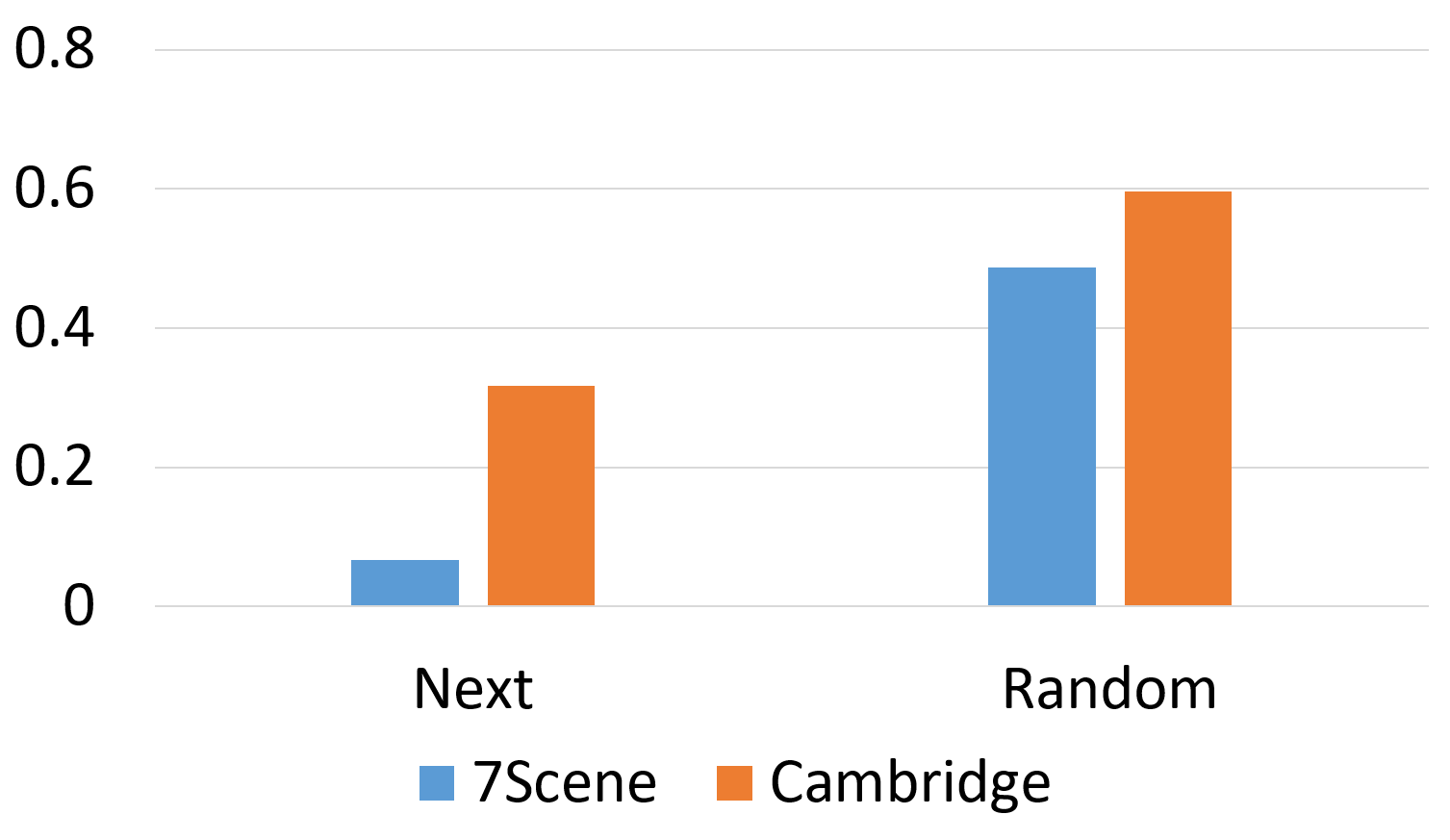}
	\caption{Average paired image similarity (measured with average Gist feature distance) on two datasets.}
	\label{fig:Sim}
\end{figure}

As shown in Table \ref{tab:refence performance:Cam}, the results of taking the next image as reference are better than that of the random reference selection strategy on both the average positional and orientational accuracy. It is also seen that randomly choosing the reference image achieves the worse performance on positional accuracy than the baseline. This may be explained by the fact that images of the \emph{Cambridge Landmarks} are of large difference so that the metric distance loss fails to work.
To verify the explanation, we measure image similarity of the two pairing strategies. The average Euclidean distance of GIST features of paired images are employed to quantify paired image similarity. The average feature distances of the scenes are shown in Figure \ref{fig:Sim}.
\begin{table*}[!ht]
	\caption{\label{tab:dataset_details} Statistic of image similarity (measured by average Gist features distance) of two image pairing strategies.}
	\centering
	\begin{tabular}{lcccccl}
		\hline
		Scene & Training  & Testing  &Spatial scope(m)&  Next &  Random \\
		\hline
		Chess & 4000 & 2000 & 3 $\times$ 2& 0.0700 & 0.5044\\
		Fire & 2000 & 2000 & 2.5 $\times$ 1& 0.0968 & 0.5187\\
		Heads & 1000 & 1000 & 2 $\times$ 0.5& 0.0613 & 0.4866\\
		Office & 6000 & 4000 & 2.5 $\times$ 2& 0.0600 & 0.4366\\
		Pumpkin& 4000 & 2000 & 2.5 $\times$ 2& 0.0540 & 0.4390\\
		Red Kitchen & 7000 & 5000 & 4 $\times$ 3& 0.0749 & 0.4993\\
		Stairs & 2000 & 1000 & 2.5 $\times$ 2& 0.0540 & 0.5220\\
		\hline
		
		KingsCollege& 1220 & 343 & 140 $\times$ 40& 0.2816 & 0.5531\\
		OldHospital& 895 & 182 & 50 $\times$ 40& 0.3338 & 0.6127\\
		ShopFacade& 231 & 103 & 35 $\times$ 25& 0.3133 & 0.5730\\
		StMarysChurch& 1487 & 530 & 80 $\times$ 60& 0.3411 & 0.6471\\
		
		\hline
	\end{tabular}
	
\end{table*}

Table \ref{tab:dataset_details} shows that for each scene, taking the next image as reference achieves higher similarity between paired images than that of randomly chosen. This confirms our explanation that MDLoss works better in scenarios where paired images are similar.
\section{Concluding Remarks} \label{sec:conclusion}

In this paper, we enhance the camera relocalization performance of deep learning-based methods by introducing the relative geometry constraints.  This is achieved by designing a relative geometry-aware  Siamese neural network and three relative geometry-related loss functions. The proposed network is capable of predicting the poses of two images as well as the relative pose between them. Another advantage of the network is that it is able to predict the global pose by feeding a single image into one stream of it. The new pose space relative loss and feature space relative regression loss functions can be combined with traditional global pose loss to enhance the position and orientation accuracy. 
The metric distance loss enables the network to learn deep feature representation that can distinguish similar images of different locations, thus helping improve localization accuracy.   
We also find that pairing similar images outperforms random paring.
In future work, we plan to investigate the combination of deep learning-based methods and 3D modeling-based methods to further enhance the performance.

\appendices

\ifCLASSOPTIONcompsoc
  \section*{Acknowledgments}
\else
  \section*{Acknowledgment}
\fi

\ifCLASSOPTIONcaptionsoff
  \newpage
\fi

\end{document}